\documentclass[runningheads]{llncs}
\usepackage[T1]{fontenc}
\usepackage{graphicx}
\usepackage{xcolor}
\usepackage{amssymb}
\usepackage{amsmath}
\usepackage{multirow}
\usepackage{url}
\usepackage{hyperref}
\usepackage[square,numbers,sort]{natbib}
\usepackage{booktabs}

\newcommand{\footurl}[1]{\footnote{\url{#1}}}

\begin{document}
%
\title{Analysing Public Transport User Sentiment on Low Resource Multilingual Data}
\titlerunning{User Sentiment on Low Resource Public Transport Data}
%

\author{Rozina Myoya\inst{1}\orcidID{0000-0002-3154-1673} \and
Vukosi Marivate\inst{1}\orcidID{0000-0002-6731-6267}\and
Idris Abdulmumin\inst{1}\orcidID{0000-0002-3795-8381}}
%
\authorrunning{Myoya et al.}
%
\institute{Department of Computer Science, University of Pretoria, Lynnwood Road, Pretoria 0002, South Africa\\
\email{rozina.myoya@tuks.co.za, \{vukosi.marivate,idris.abdulmumin\}@up.ac.za}\\ 
}

%
\maketitle              
\begin{abstract}
Public transport systems in many Sub-Saharan countries often receive less attention compared to other sectors, underscoring the need for innovative solutions to improve the Quality of Service (QoS) and overall user experience. This study explored commuter opinion mining to understand sentiments toward existing public transport systems in Kenya, Tanzania, and South Africa. We used a qualitative research design, analysing data from X (formerly Twitter) to assess sentiments across rail, mini-bus taxis, and buses. By leveraging Multilingual Opinion Mining techniques, we addressed the linguistic diversity and code-switching present in our dataset, thus demonstrating the application of Natural Language Processing (NLP) in extracting insights from under-resourced languages.  We employed PLMs such as AfriBERTa, AfroXLMR, AfroLM, and PuoBERTa to conduct the sentiment analysis. The results revealed predominantly negative sentiments in South Africa and Kenya, while the Tanzanian dataset showed mainly positive sentiments due to the advertising nature of the tweets. Furthermore, feature extraction using the Word2Vec model and K-Means clustering illuminated semantic relationships and primary themes found within the different datasets. By prioritising the analysis of user experiences and sentiments, this research paves the way for developing more responsive, user-centered public transport systems in Sub-Saharan countries, contributing to the broader goal of improving urban mobility and sustainability.

\end{abstract}%
\section{Introduction}\label{section1}

Mapping the user experience of public transportation, which is an integral part of millions of commuters daily, provides the opportunity to illuminate user behavior and identify areas of improvement that when implemented could boost public transport usage. Social media has become a rich source for gauging public sentiments, with commuters frequently and openly sharing feedback, thus creating an opportunity for transit authorities to leverage social media to enhance user interaction and the user experience \cite{new_technology,cndro_2021, haghighi2018using}. For instance, \cite{cndro_2021} noted that transit authorities use social media as a way of engaging with the public, providing timely updates, alerting the public about emergencies, and promoting the usage of public transportation. 

This study explored the application of NLP in a multilingual context to gain insight into the commuter experience by carrying out sentiment analysis on the data extracted from X (formerly Twitter). Given the expected multilingual nature of the data we used existing annotated datasets that included the following languages: English, Swahili, SeTswana and isiZulu to carry out the testing and evaluation of the chosen Pre-trained Language Models (PLMs) that would be used to conduct the sentiment analysis \citep{mabokela2022sentiment, mambina2022classifying}.



This research aimed to showcase the practical application of multilingual sentiment analysis for in-depth market insights in public transport, with the aim of developing short to medium-term solutions to enhance the user experience. 

The main contributions of this paper are:
\begin{enumerate}
    \item An in-depth analysis of user sentiments regarding public transport in Sub-Saharan countries using data from X.
    \item Evaluation of experiences across major transport modes such as rail, mini-bus taxis, and buses.
    \item Application of Multilingual Opinion mining to handle language diversity and code-switching in the dataset.
    \item Demonstration of the practical use of NLP in extracting insights from under-resourced language data to strategically enhance public transport systems.
    
\end{enumerate}


\section{Related work}\label{section2}

Traditionally, user experiences in public transport systems have been studied through extensive and resource-intensive methods such as surveys and observational studies, however, the rise of social media platforms like X offers dynamic, real-time avenues for collecting similar user experiences \cite{casas2017tweeting, keseru2020multitasking, rieser2013identifying}. 

In recent years, social media has revolutionised opinion mining platforms such as WhatsApp, Facebook, and X present opportunities to capture diverse opinions and sentiments across demographic and geographic segments \cite{statista_2023}, particularly in the public transport sector. However, using private channels like WhatsApp raises concerns about data privacy and consent, making platforms like Facebook and X more viable for user experience analysis. This shift has also seen transit agencies increasingly utilise social media for public engagement and service improvement, as exemplified by campaigns like South Africa's \#adoptastation \cite{alexander2012scotrail,cndro_2021}.
This shift however also introduces challenges in analysing social media data, which often includes colloquial, noisy, and code-mixed content \cite{ledwaba2022semi}, thus necessitating high-quality annotated datasets for effective sentiment analysis.

The rise of social media opinion mining has also spurred the demand for advanced Natural Language Processing (NLP) tools, especially in sentiment analysis, which is crucial for understanding user experiences in multilingual and low-resource language contexts \cite{qi2019study}. Tools such as AfroLID \cite{adebara2022afrolid}, Franc, and CLD3 \cite{salcianu2018compact} have emerged to address the challenges of language identification in opinion mining. Furthermore, selecting the appropriate language model, whether developed from scratch or fine-tuned from pre-trained models, is crucial for accurate analysis in a multilingual context \citep{ekinci2018segmentation, kuratov2019adaptation}. 

The emergence of pre-trained language models (PLMs) has revolutionised natural language processing (NLP) by enabling complex linguistic tasks to be handled with remarkable precision and efficiency \cite{wang2022pre}. However, a significant limitation arises when these models are applied to African languages, which are often underrepresented in global datasets. This has led to the development of innovative solutions such as multilingual adaptive fine-tuning, which extends the capabilities of PLMs to low-resource languages \cite{alabi2022adapting, afrolm, ogueji2021small}. Despite the effectiveness of this approach, it requires substantial resources, highlighting the need for models pre-trained specifically on target languages \cite{alabi2022adapting}. In response, specialised models such as AfriBERTa \cite{ogueji2021small}, AfroXLMR \cite{alabi2022adapting}, AfroLM \cite{afrolm}, and PuoBERTa \cite{marivatePuoBERTa2023} have been developed, offering enhanced accessibility and applicability for African languages in critical NLP tasks like sentiment analysis.

In Nairobi, public transport accounts for 36.4\% of commuter trips, with a significant portion of the population walking or cycling due to the high cost of bus tickets and severe traffic congestion deterring private car usage \cite{githui2009structure}. Similarly, in Dar es Salaam, the growing demand for public transport has led to the dominance of minibuses and the implementation of the Bus Rapid Transit system, reflecting the city's adaptive response to its expanding urban needs \cite{kanyama2004public, kruger2021bus}. In South Africa, despite efforts to shift commuters from private cars to public transport to reduce congestion and pollution, private car usage continues to rise, indicating that public transport is often used out of necessity rather than choice \cite{clark2002public, vicente2016profiling}. This scenario presents a significant challenge for transport providers in sub-Saharan Africa, where public transport investment is often de-prioritised. This study aimed to address these challenges by mining commuter sentiment to better understand their experiences and preferences, potentially guiding improvements in the respective public transport systems.

\section{Methodology}\label{section3}
This study used data extracted from X from three different countries, namely, Kenya, South Africa, and Tanzania. The data was multilingual with a mixture of English, Swahili, SeTswana, and isiZulu, therefore consideration was taken in analysing code-mixed data and choosing machine learning models that were pre-trained on these respective languages. The methodology for data extraction, processing, and analysis were structured into five steps: (i) Data sourcing; (ii) Data pre-processing and Exploratory Data Analysis; (iii) Training data description and analysis; and (iv) Model testing and evaluation; 

\subsection{Data Sourcing}\label{section3.1}
The keywords pertinent to public transport within each country used in the data extraction process are presented in Table~\ref{tab:keywords}. The data timeline was between $1^{st}$ of Jan 2007 and $1^{st}$ of March 2023. 

\begin{table}[!ht]
\caption{Transport keywords for each country}
\label{tab:keywords}
\begin{tabular}{ll}
\toprule
\textbf{Country}  & \textbf{Transport keywords} \\ \midrule
Kenya        &  `matatu', `kencom', `citihoppa', `public transport', `boda boda', \\
             &`bus', `kbs', `mathree', `ma3', `tuk tuk', `BasiGo', `hoppaciti'     \\ \midrule
Tanzania     &  `daladala', `boda boda', `dart', `bus', `train', `ferry', `bajaj'   \\ \midrule
South Africa &   `prasa', `metrorail', `taxi', `public transport', `putco', `gautrain',\\
             &`rea vaya', `a re yeng', `bus', `train', `shosholoza meyl', `metrobus'    \\ \bottomrule
\end{tabular}%
\end{table}

To ensure relevance, the data extraction was confined to major metropolitan areas in each country: Nairobi, Dar es Salaam, and Johannesburg. These cities, being the nerve centers of public transport, provided a comprehensive view of commuter sentiments. The total size of the extracted dataset for each country is presented in Table~\ref{tab:dataset_size}. 

\begin{table}
    \caption{Extracted dataset sizes}
    \centering
    \begin{tabular}{lr}
    \toprule
     \textbf{Country}    & \textbf{No. of tweets}\\  \midrule
      Kenya        &  5,887\\
      Tanzania     &  998\\
      South Africa &   7,661 \\  \midrule
      Total dataset size &  14,546 \\  \bottomrule
    \end{tabular}
    \label{tab:dataset_size}
\end{table}

\subsection{Data Processing and Exploratory Data Analysis (EDA)}\label{section3.2}

Data processing included the removal of punctuation, trimming of spaces, and expanding contractions (e.g., "what's" to "what is"). Transport-related keywords, initially used for data sourcing, were discarded. We also removed English and Swahili stop words and named entities to emphasise the data's primary features. Furthermore, to protect user privacy, all usernames and location tags were discarded from the dataset. The EDA process also involved Language Identification (LID) and feature extraction that was used to understand the semantic interplay among words.

\subsubsection{Language Identification (LID)}
To determine the language distribution in each dataset and filter relevant languages for sentiment analysis, we conducted a Language Identification (LID) task using three tools: \textit{AfroLID} \citep{adebara2022afrolid}, \textit{Franc}, and \textit{CLD3} \citep{salcianu2018compact}. The choice of these tools was influenced by the findings of \cite{adebara2022afrolid}, which highlighted their superior performance in detecting African languages. After initial identification, manual validation was performed on a 10\% sample from each dataset, categorised by source country. The most accurate LID tool was then selected for further tasks. The performance comparison of each model, based on manual validation, is provided in Table~\ref{tab:lid_validation}. On average, Franc outperformed CLD3 and AfroLID in the validation process, proving to be the most reliable in terms of result consistency and its ability to identified all the languages under consideration. Therefore, Franc was chosen for subsequent steps involving identifying code-mixed sentences within the dataset.

\begin{table}[!ht]
\centering
\caption{LID accuracy scores after result validation}
\label{tab:lid_validation}
\begin{tabular}{llll}
\toprule
\textbf{Language} & \textbf{AfroLID} & \textbf{CLD3} & \textbf{Franc} \\ \midrule
Swahili           & 48\%             & 84\%          & 73\%          \\
SeTswana          & 50\%             & -             & 75\%          \\
isiZulu           & 33\%             & -             & 44\%          \\
English           & -                & 60\%          & 46\%          \\ \bottomrule
\end{tabular}
\end{table}
 
\begin{figure}[!ht]
    \centering
    \includegraphics[width=0.9\textwidth]{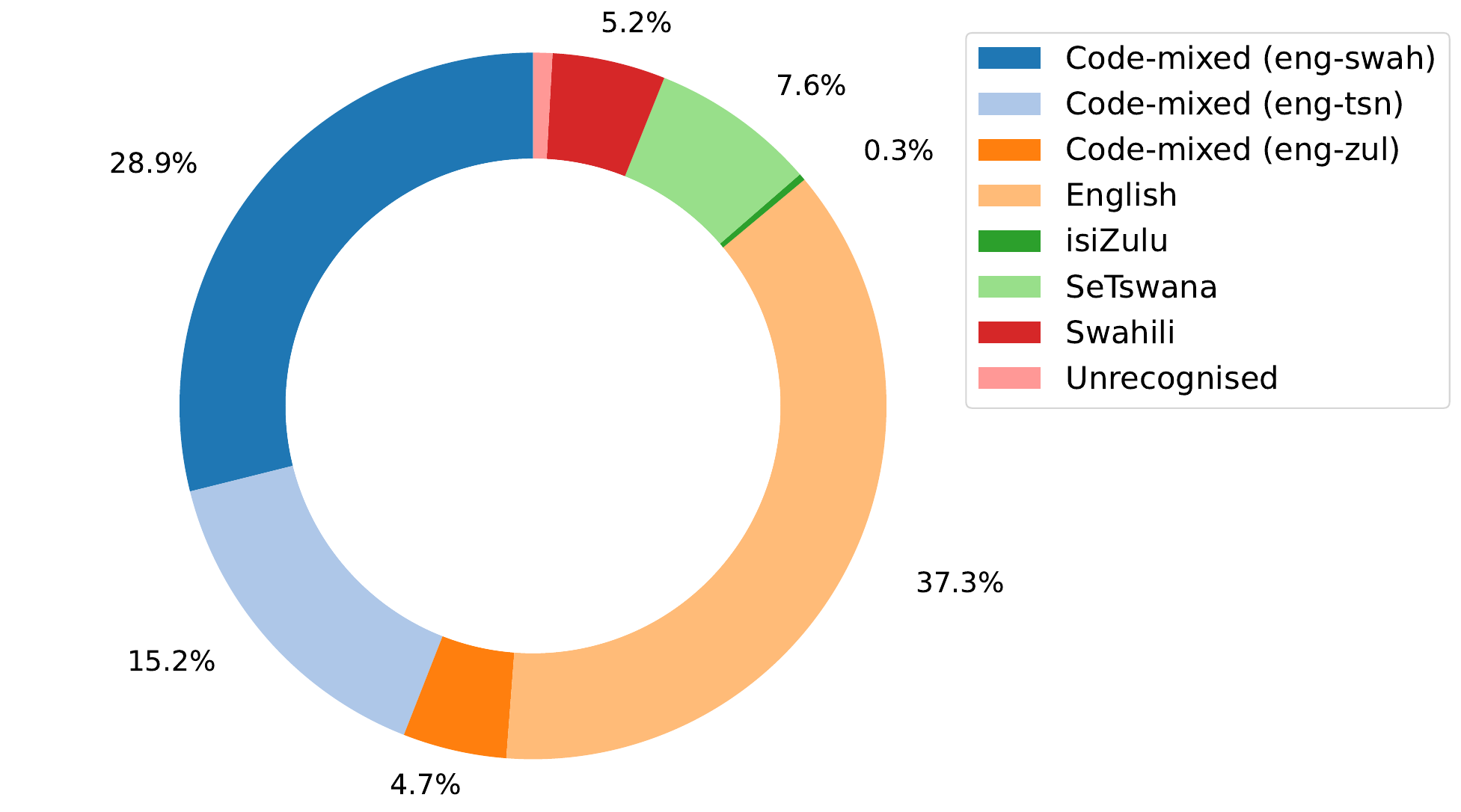}
    \caption{Focus language distribution within the full dataset taking into consideration code mixing}
    \label{fig:code_mixed_lang_dist}
\end{figure}

Code-mix identification involved tokenizing each sentence within the dataset and applying LID using Franc on a word-by-word basis. The languages identified within each sentence were then cataloged to ascertain the extent of code-mixing. We first narrowed down the dataset using the initial LID results from the Franc model, selecting sentences identified as one of the focus languages. This filtration step adjusted the dataset to the size of 12 817 tweets. Following this,
we applied the word-by-word LID method to pinpoint code-mixed sentences, with the findings presented in Figure~\ref{fig:code_mixed_lang_dist}.

\subsubsection{Feature extraction}
\begin{figure}[!ht]
    \centering
    \includegraphics[width=0.85\columnwidth]{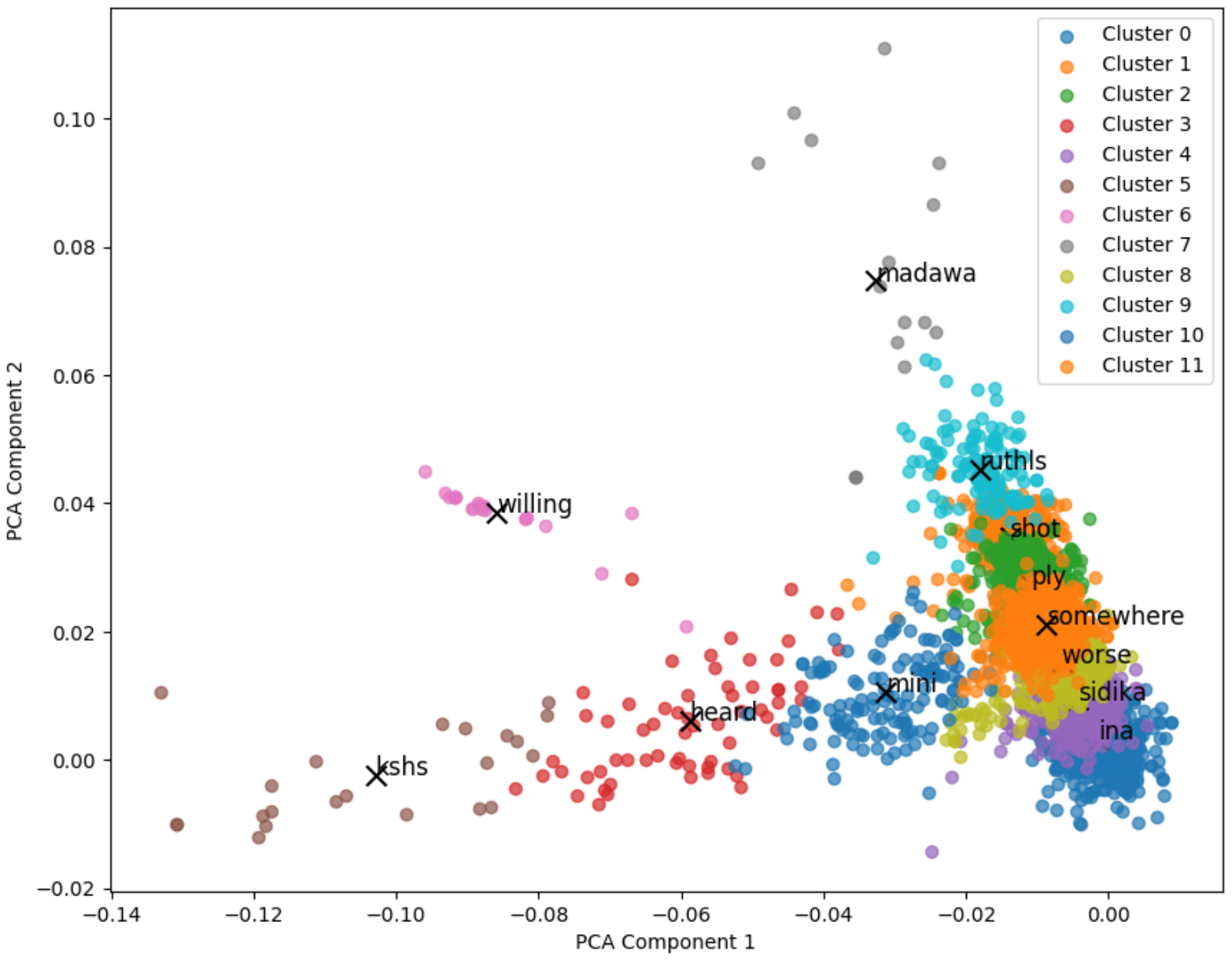}
    \caption{Main features extracted from the Kenyan dataset}
    \label{fig:ken_main_features}
\end{figure}

\begin{figure}[!ht]
    \centering
    \includegraphics[width=0.85\columnwidth]{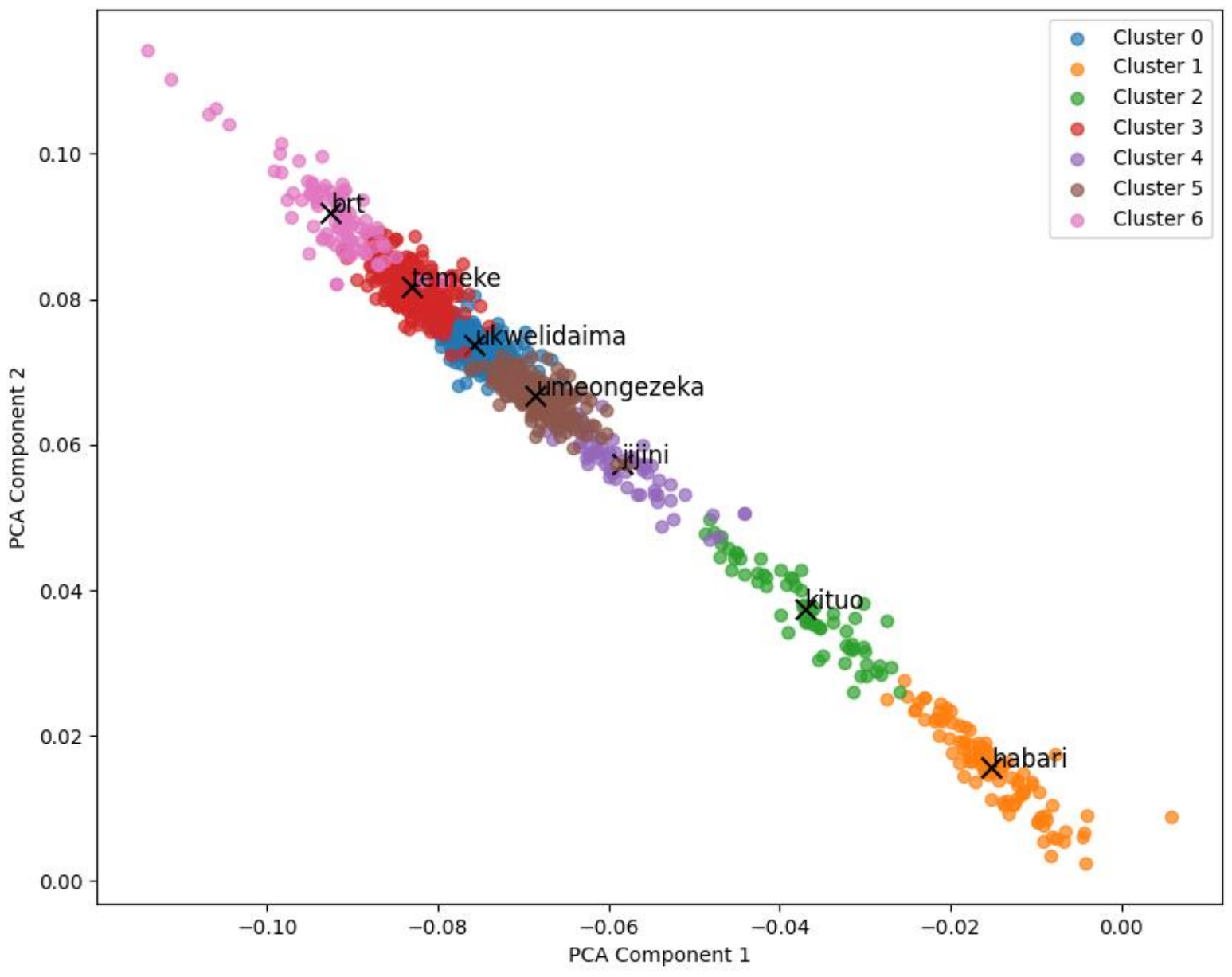}
    \caption{Main features extracted from the Tanzanian dataset}
    \label{fig:tz_main_features}
\end{figure}

\begin{figure}[!ht]
    \centering
    \includegraphics[width=0.85\columnwidth]{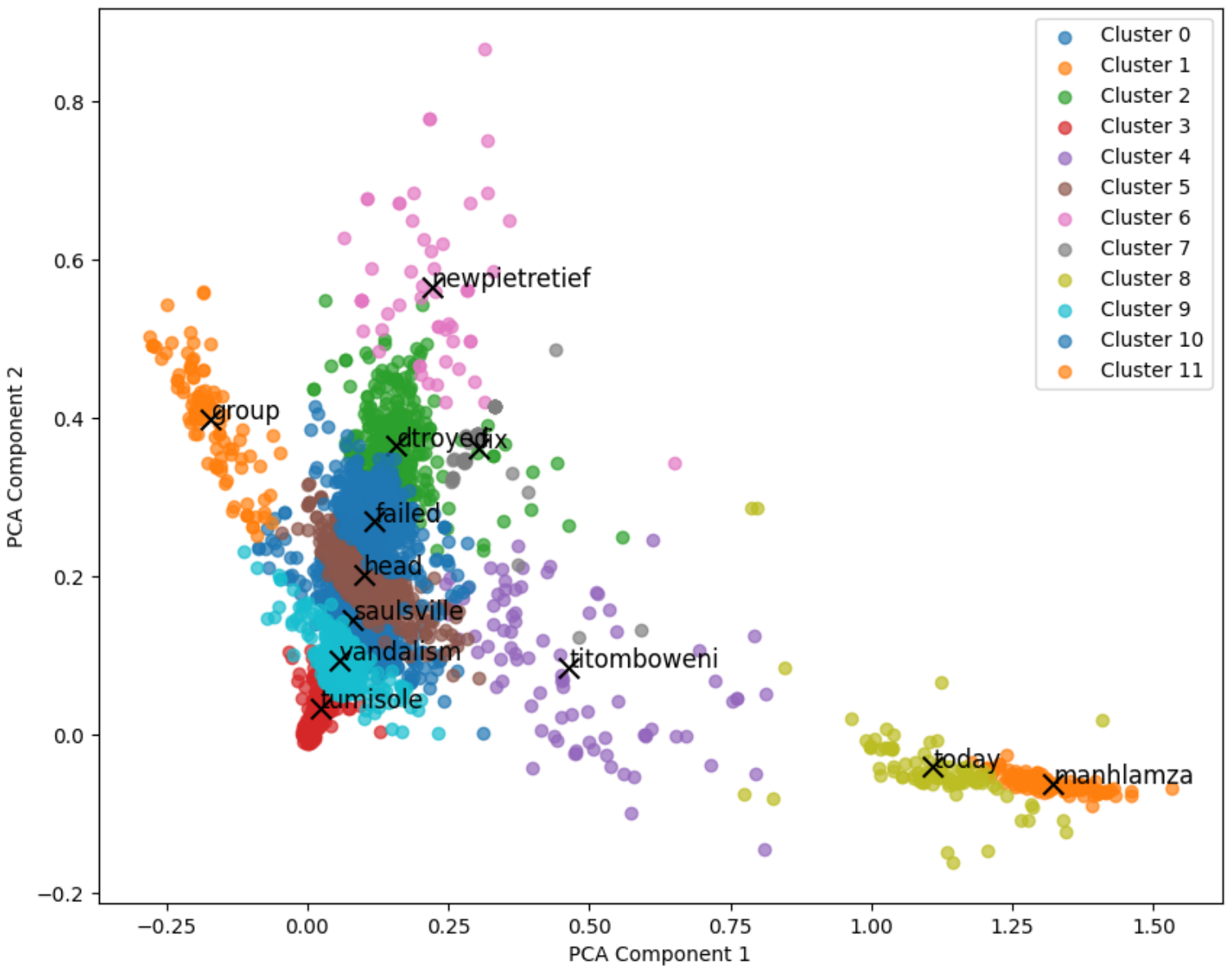}
    \caption{Main features extracted from the South African dataset}
    \label{fig:sa_main_features}
\end{figure}

Feature extraction was conducted by determining word embeddings derived using
the Word2Vec model. The Word2Vec model was trained on each tweet containing a minimum of 5 words and maximum of 20 words, using a vector size of 100 and a default window size of 5. This configuration strikes a balance between capturing local syntactic information and broader semantic context\citep{tensorflow_word2vec}. Subsequent clustering (K-Means clustering) was performed to discern the semantic relationships between these words. Each country's dataset underwent this process, and the results revealed that topic modeling could be applied to each dataset, and themes could be inferred from the clusters of the primary features extracted from the data. The results are presented in Figures~\ref{fig:ken_main_features} to \ref{fig:sa_main_features}.


To highlight some of the main themes derived from the Kenyan dataset, the term ``kshs" was linked to public transport pricing, such as bus and matatu tickets, indicating the effect of pricing on commuter sentiment. ``Ruthls" touched on sentiments towards police, military, and government being ruthless towards its citizens and thus indicating negative sentiment. ``Madawa" (Swahili for ``drugs") highlighted concerns about passenger drugging in \textit{matatu} travel and highlighted safety concerns of commuters, emphasised also by the term ``Shot". Thus highlighting two of the main concerns within the Kenyan commuter population being the cost of public transport and safety. 

Shifting focus to Tanzania, the ``brt" feature predominantly revolved around the Bus Rapid Transit (BRT) system in Dar es Salaam. ``Umeongezeka," translating to ``has increased" in Swahili, referred to fare hikes and thus the significant effect of pricing on commuter sentiment. Features such as ``Jijini" and ``kituo" brought the city and station into focus with comments on the quality of infrastructure within the city and bus stations, thus indicating the effect of city and station conditions on commuter sentiment. In summary, this highlighted the two of the main concerns within the Tanzanian commuter population also being cost of public transport, with the addition for the concern of the quality of infrastructure within the system. 

The main themes derived from the South African dataset were concerns for the deteriorating state of the public transport system that were highlighted by terms such as ``Destroyed" and ``failed". ``Vandalism" encapsulated narratives around damage to the transport infrastructure and stations, particularly within the rail system, while the term``titomboweni" resonated with discussions around Tito Mboweni, the then Finance Minister, and his association with PRASA's challenges such as state capture and corruption. The overarching theme derived from this dataset showed that some of the main user concerns fall within infrastructure quality concerns, and the government's role in public transport provision.

\subsection{Training Dataset Description}\label{section3.3}
The training datasets were a compilation of tweets labeled to reflect the three traditional sentiment categories: positive, neutral, and negative. The size and distribution of the sentiments of the training datasets used in the study are summarised in Table~\ref{tab:training_datasets}. These annotated datasets were sourced from \cite{muhammad-etal-2023-afrisenti} and the Data Science for Social Impact (DSFSI)\footnote{https://dsfsi.github.io/} research group from the University of Pretoria.

\begin{table}[!ht]
\centering
\caption{Training datasets}
\label{tab:training_datasets}
\begin{tabular}{lllllr}
\toprule
\textbf{Dataset} & \textbf{Sources} & \textbf{Languages} & \textbf{Quality} & \textbf{Labels} & \textbf{\# Tweets} \\ \midrule
\multirow{3}{*}{\textit{Afrisenti}} & \multirow{3}{*}{\cite{muhammad-etal-2023-afrisenti}} & \multirow{3}{*}{Swahili} & \multirow{3}{*}{High} & Neutral & 1,790 \\
&&&& Positive & 1,235 \\
&&&& Negative & 387 \\ \midrule
\multirow{3}{*}{SeTswana} & \multirow{3}{*}{DSFSI} & \multirow{3}{*}{SeTswana} & \multirow{3}{*}{Medium} & Positive & 96 \\
&&&& Negative & 54 \\
&&&& Neutral & 19 \\ \midrule
\multirow{3}{*}{isiZulu} & \multirow{3}{*}{DSFSI} & \multirow{3}{*}{isiZulu} & \multirow{3}{*}{Medium} & Negative & 110 \\
&&&& Positive & 53 \\
&&&& Neutral & 16 \\ \midrule
\multirow{4}{*}{Code-mixed} & \multirow{4}{*}{DSFSI} & English, & \multirow{4}{*}{Low} & Positive & 509 \\
&& Swahili, && Negative & 242 \\
&& isiZulu, && Neutral & 39 \\
&& SeTswana && \\ \bottomrule
\end{tabular}
\end{table}

SMOTE \cite{chawla2002smote} was used to handle the class imbalance in the different datasets as shown in Table~\ref{tab:training_datasets}. This process involved tokennizing the tweets within the datasets and extracting the embeddings using a Word2Vec model. The word embeddings were then used to generate the synthetic sample.

\subsection{Model Testing and Evaluation}\label{section3.4}
The NVIDIA Tesla T4 GPU was used for model testing and evaluation. The GPU is based on a Turing architecture and comes with 16 GB GDDR6 VRAM. PLMs were used in the study to conduct the sentiment classification task. The chosen PLMs were pre-trained on the languages included in our dataset and testing was conducted taking into account the different languages present within the dataset. Model details are presented in Table~\ref{tab:model_properties}. The hyperparameters for the models were obtained through a fine-tuning process involving iterative experimentation and optimisation techniques, such as grid search. The final hyperparameters are presented in Table~\ref{tab:models_hyperparameters}. After testing and evaluating the models on the annotated datasets, using the aforementioned hyperparameters, sentiment classification on the raw data was carried out by selecting the model with the highest F1 score. The respective F1 scores are detailed in Table~\ref{tab:dataset_results}. The best-performing models for each language, based on the above performance criteria, were \textit{AfriBERTa} for Swahili, \textit{AfroXLMR-base} for isiZulu and the code-mixed dataset, and \textit{AfroLM} for SeTswana.

\begin{table}[!ht]
    \caption{Summary of the model details}
    \centering
    \begin{tabular}{lllr}
    \toprule
     \textbf{Model} & \textbf{Language included} & \textbf{Language trained} & \textbf{Size}  \\ \midrule
    AfriBERTa \citep{ogueji2021small} & Swahili & Swahili & 126M  \\ 
    AfroXLMR \citep{alabi2022adapting} & Swahili, isiZulu, English& Swahili, isiZulu, and code-mixed & 278M  \\ 
    AfroLM \citep{afrolm} & Swahili, isiZulu, Setswana& Swahili, isiZulu, and Setswana & 264M  \\ 
    PuoBERTa\citep{marivatePuoBERTa2023} & Setswana & Setswana & 83.5M    \\ \bottomrule
    \end{tabular}
    \label{tab:model_properties}
\end{table}

\begin{table}[h!]
\centering
\caption{PLM Hyperparameters}
\label{tab:models_hyperparameters}
\begin{tabular}{lllll}
\hline
Hyperparameter              & AfriBERTA   & AfroXLMR     & AfroLM       & PuoBERTa \\ \hline
Per Device Train Batch Size & 16          & 16           &16            & 16 \\
Per Device Eval Batch Size  & 16          & 16           &16            & 16\\
Epochs                      & 2           & 10           &10            & 2\\
Weight Decay                & 0.02839     & 0.2249       &0.0494        & 0.02839  \\
Seed                        & 16978       & 18790        &14640         & 16978\\
Learning Rate               & 5e-5        & 4e-5         &5e-5          & 5e-5\\
Adafactor                   & True        & True         & True         & True\\
Adam $\beta_1$              & 0.7640      & 0.2727       & 0.5542       & 0.7640\\
Adam $\beta_2$              & 0.7439      & 0.6734       &0.9000        & 0.7439  \\
Adam $\epsilon$             & 3e-8        & 3e-10        &3e-8          & 3e-8\\
Max Gradient Norm           & 0.4773      & 0.8040       &0.1338        &  0.4773 \\
Metric for best model       & `eval\_loss'& `eval\_loss' & `eval\_loss' & `eval\_loss' \\
Gradient Accumulation Steps & 1           & 1            &1             &1\\
Warm up steps               & 0           & 0            &0             &0\\
Dataloader num of workers   & 4           & 4            & 4            &4 \\ \hline
\end{tabular}
\end{table}

\begin{table}[!ht]
\caption{Model Evaluation (F1-Score) }
\centering
\resizebox{\textwidth}{!}{
\begin{tabular}{lrrrr}
\toprule
\textbf{Model} & \multicolumn{1}{l}{\textbf{Swahili (\texttt{swh})}} & \multicolumn{1}{l}{\textbf{isiZulu (\texttt{zul})}} & \multicolumn{1}{l}{\textbf{Setswana (\texttt{tsn})}} & \multicolumn{1}{l}{\textbf{Code-mixed (with \texttt{eng})}} \\ \midrule
AfriBERTa       & 0.61                        & -                           & -                            &                       -         \\
AfroXLMR (base) & 0.59                        & 0.82                        & -                            &         0.97                       \\
AfroLM          & 0.58           &    0.59   &   0.62                        &                      -          \\
PuoBERTa        & -                           & -                           & 0.43                         & -                              \\ \bottomrule
\end{tabular}
}
    \label{tab:dataset_results}
\end{table}

 
\newpage
\section{Sentiment Analysis Results}\label{section4}
The sentiment classification was carried out by segmenting the sourced data into the different themes derived from the feature extraction process in Section~\ref{section3.2}. 

\subsubsection{Kenya}
\begin{figure}[!ht]
    \centering
    \includegraphics[width=\columnwidth]{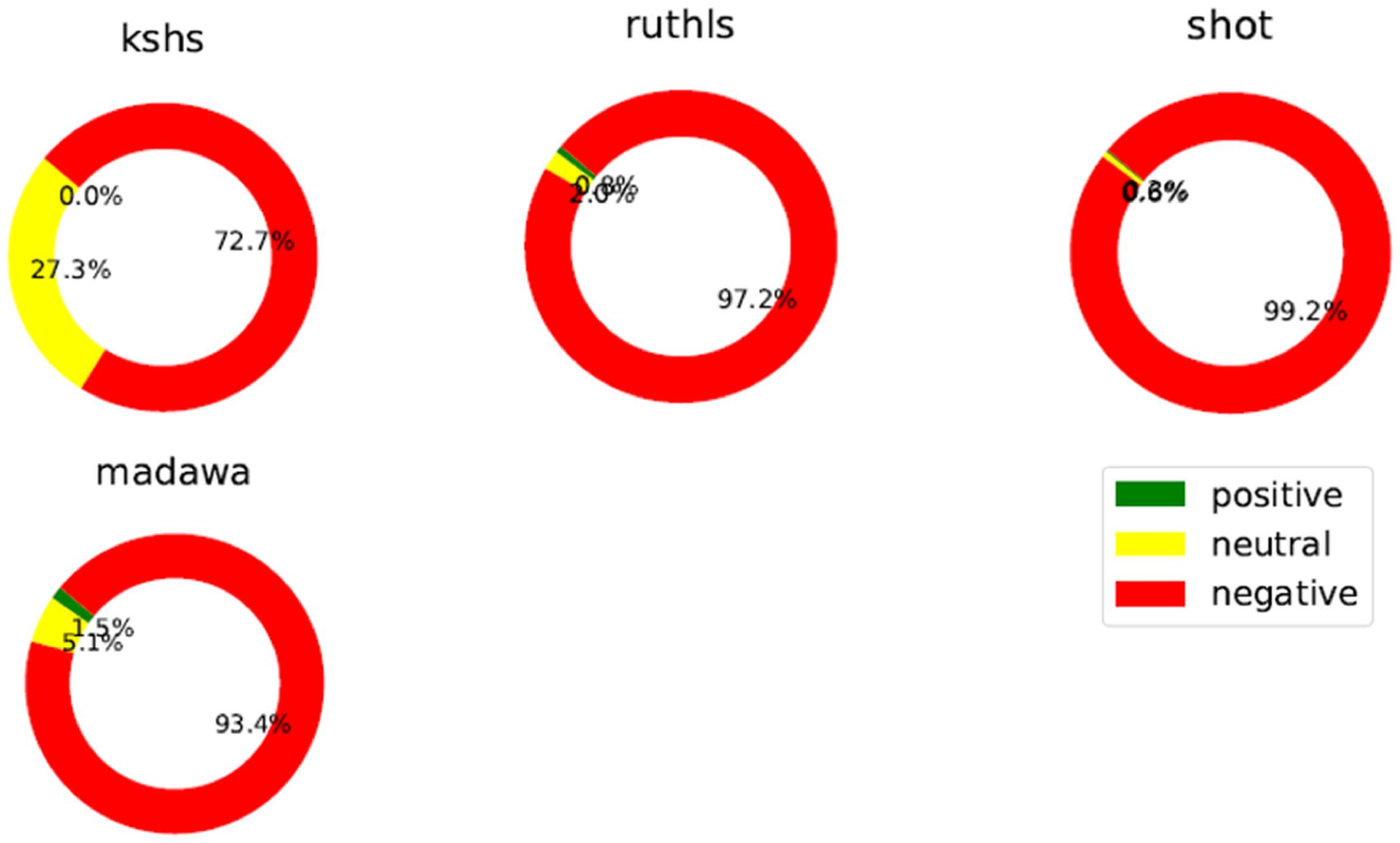}
    \caption{Sentiment distribution of Kenyan tweets according to the themes derived from Section \ref{section3.2}}
    \label{fig:theme_sentiment_ken}
\end{figure}

The results showed that within the Kenyan dataset, the predominant sentiment was negative, echoing the safety concerns of commuters. The primary themes extracted included unpredictable price hikes, over-speeding, violent crime, and safety concerns. Within each of these themes the sentiment was found to be predominantly negative as shown in Figure~\ref{fig:theme_sentiment_ken}. Further analysis of the data points within each theme suggested that these issues were primarily centred around \textit{Matatu} usage. Since 2004, the \textit{Matatu} industry, which dominates the minibus taxi sector, has been the subject of various reform initiatives aimed at improving safety and operational standards \cite{githui2009structure}. These reforms sought to address issues such as reducing speeding-related accidents, enhancing commuter safety, ensuring driver and conductor competence, removing unauthorised personnel, and regulating vehicle operations \cite{chitere2004efforts}. However, despite strict enforcement of the Traffic Act 403, \textit{Matatus} remain a leading contributor to road accidents \cite{mburu2023developing}. Additionally, the industry has been plagued by incidents of commuter harassment, particularly affecting women \cite{mwaura2020making}. Employing opinion mining for incident detection could provide near real-time insights and lead to more timely, actionable solutions.

\subsubsection{Tanzania}
\begin{figure}[!ht]
    \centering
    \includegraphics[width=\columnwidth]{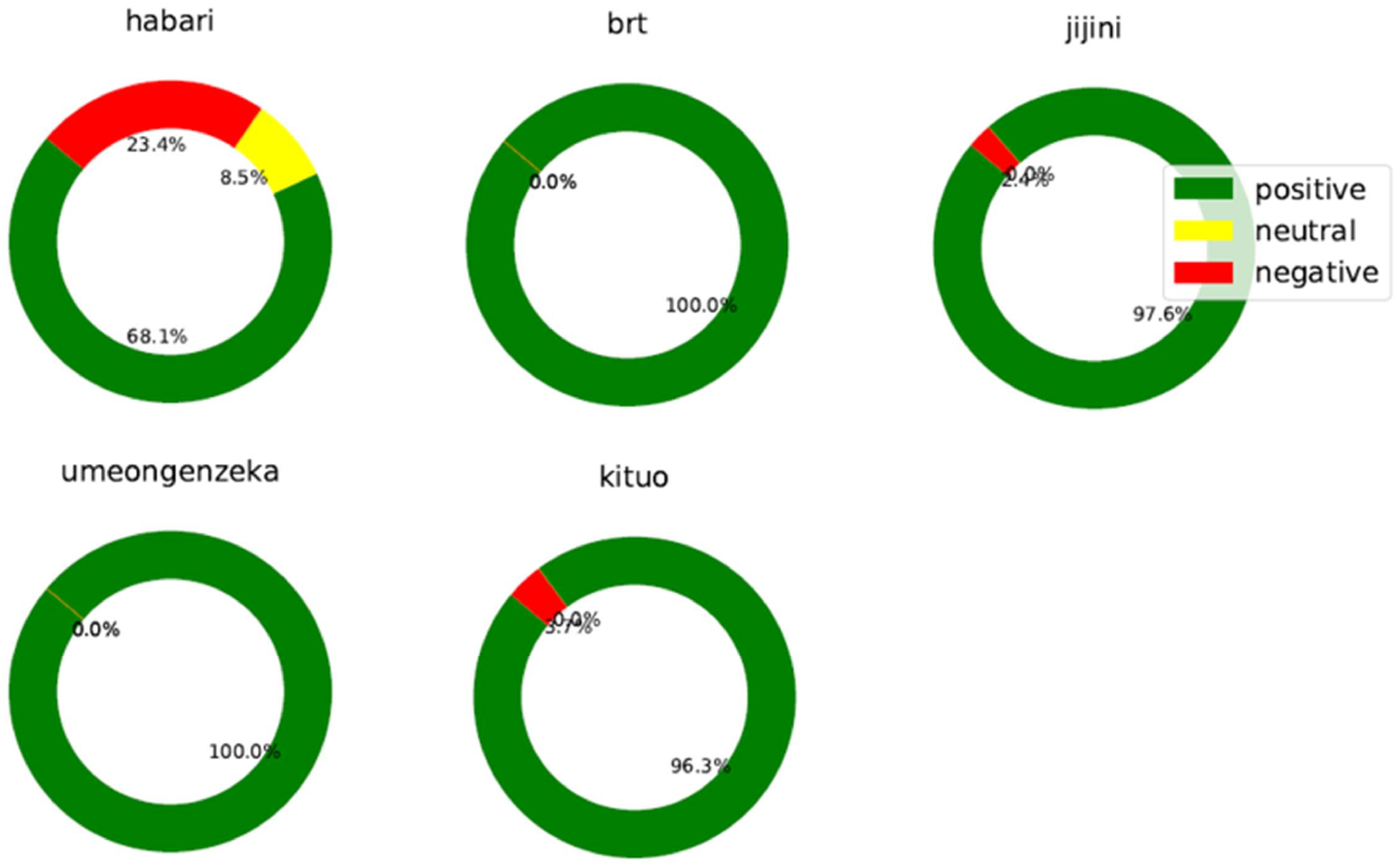}
    \caption{Sentiment distribution of Tanzanian tweets according to the themes derived from Section \ref{section3.2}}
    \label{fig:theme_sentiment_tz}
\end{figure}

The analysis of the Tanzanian dataset revealed a predominantly positive sentiment across the thematic subsets, as illustrated in Figure \ref{fig:theme_sentiment_tz}, except for the theme `habari'. This particular theme, which primarily includes tweets related to news reporting, accounted for the presence of a few negative tweets. The overall positive bias identified within these themes can be attributed to the nature of the tweets in the Tanzanian dataset, which were largely focused on advertising. Therefore, in future a more representative dataset of the Tanzanian commuter experience would be required. This also highlights that the use of X to derive commuter sentiment within the Tanzanian context may not be effective. 

\subsubsection{South Africa}
\begin{figure}[!ht]
    \centering
    \includegraphics[width=\columnwidth]{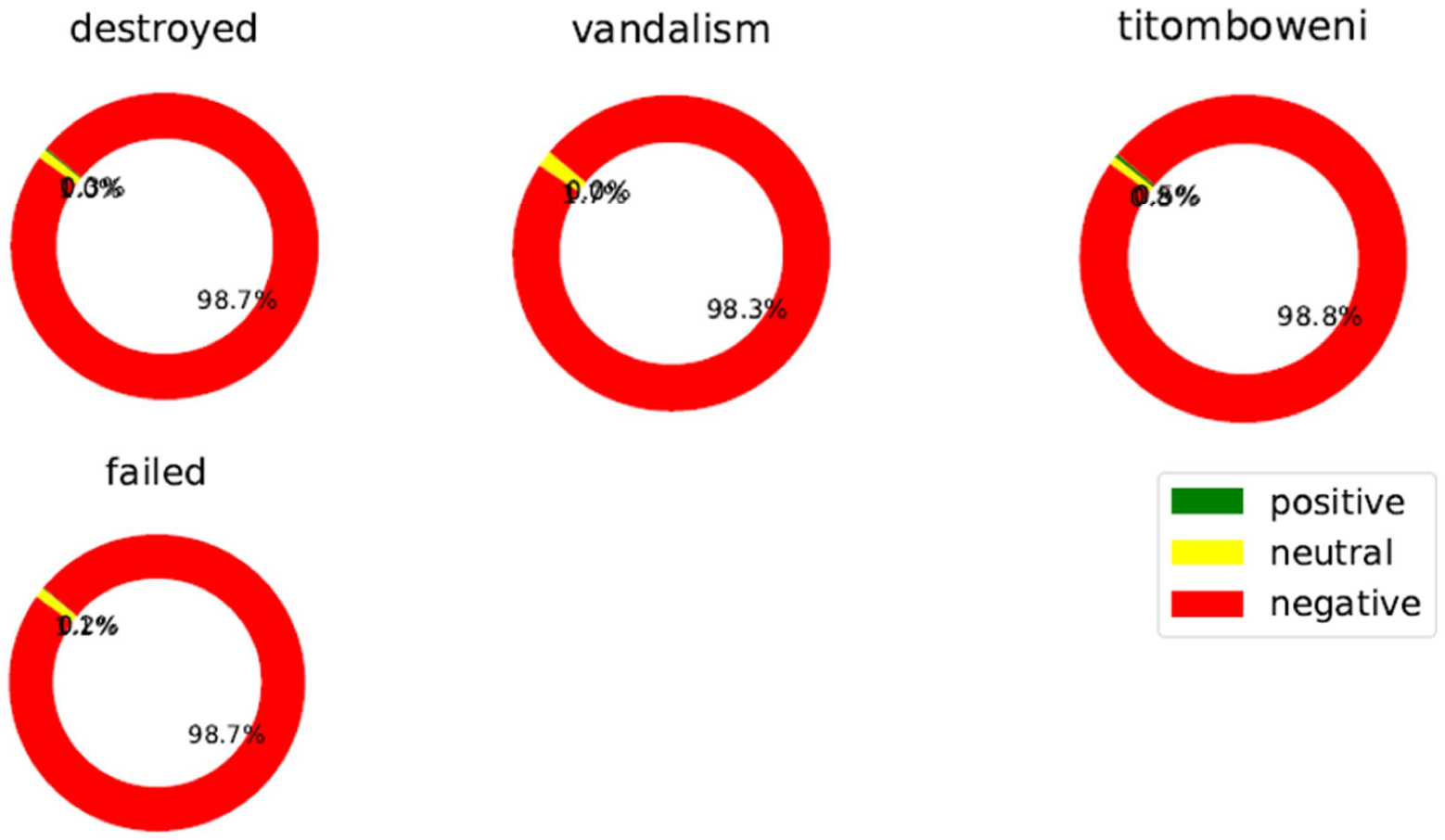}
    \caption{Sentiment distribution of South African tweets according to the themes derived from Section \ref{section3.2}}
    \label{fig:theme_sentiment_sa}
\end{figure}

Within the South African dataset, the findings illustrated in Figure \ref{fig:theme_sentiment_sa} revealed a predominantly negative sentiment across the theme subsets. This negativity was associated with themes identified during the feature extraction process, particularly those related to destruction, failure, and vandalism. Further examination of the data points within each theme indicated that these issues were largely focused on train usage. In South Africa, the Railway Safety Regulator (RSR) annually reports on railway safety, adhering to the South African National Standards (SANS) categories \cite{sans3000-1:2009}. The sentiment expressed by commuters in tweets related to these themes aligns with RSR reports, which show a steady increase in the ratio of security-related incidents to operational occurrences since 2017—a period that corresponds with the timeline of the extracted tweets \citep{railway2021safety}. Furthermore, with the decline of the PRASA Metrorail service (train service), there is growing concern about under-reporting, making the safety data less reliable each year. To bridge this information gap between service providers and commuters, supplementary data sources like the opinion mining presented in this study can be invaluable


\section{Conclusions and Future Work}\label{section5}
The study demonstrated the successful application of NLP techniques to analyse public transport user sentiment. This approach highlights the potential of leveraging social media data to gain insights into user experiences and service gaps in public transport systems, particularly in Sub-Saharan Africa. The results derived from the feature extraction process coupled by the results from the sentiment analysis show a predominantly negative sentiment towards public transport with the overarching commuter concerns being the cost of public transport usage, safety, the state of infrastructure, and the government's role in public transport provision. 

Looking ahead, we propose expanding the range of raw data collected and implementing thorough validation processes. These raw datasets should align with the themes extracted from our dataset and the metrics used by transport providers, with the aim of facilitating more comprehensive and actionable insights. Furthermore, incorporation of methods such as aspect-based opinion mining could be used to to gain deeper insights into commuter sentiments regarding specific features of the public transport system. By integrating these strategies, we not only aim to improve the quality of commuter experiences but also aspire to establish a benchmark for responsible and impactful NLP applications within the transportation sector.




\subsubsection{Limitations}
The models' generalisability across diverse contexts was a challenge, especially given the linguistic diversity and code-mixing prevalent in regions like Sub-Saharan Africa. Additionally, the dynamic nature of social media language necessitates regular model updates to remain relevant and accurate. Furthermore, pre-trained models, though advantageous, might not always capture regional linguistic nuances.

\subsubsection{Ethical Considerations}
In our research, we sourced data from a social media platform that may contain users' personal details, therefore we prioritised the privacy and confidentiality of these users by meticulously removing any identifiable information, including usernames and location tags. 




\subsubsection{Acknowledgments}
This research was conducted by the Computer Science Department at the University of Pretoria under the Data Science for Social Impact (DSFSI) research group. We would therefore like to express our deepest gratitude to the members of the DSFSI research group for providing GPU resources and annotated datasets to conduct the research. 

%

\bibliographystyle{abbrvnat}
\bibliography{reference}
\end{document}